\documentclass[conference]{IEEEtran}

\usepackage{float}
\usepackage{amssymb}
\usepackage{balance}
\usepackage{color}
\usepackage{multirow}
\usepackage{cite}
\usepackage[numbers]{natbib}

\begin{document}
%
\title{Adversarial Analysis of Natural Language Inference Systems}

\author{\IEEEauthorblockN{Tiffany Chien}
\IEEEauthorblockA{University of California, Berkeley}
\and
\IEEEauthorblockN{Jugal Kalita}
\IEEEauthorblockA{University of Colorado, Colorado Springs}}


%


\maketitle

\begin{abstract}
The release of large natural language inference (NLI) datasets like SNLI and MNLI have led to rapid development and improvement of completely neural systems for the task.
Most recently, heavily pre-trained, Transformer-based models like BERT and MT-DNN have reached near-human performance on these datasets.
However, these standard datasets have been shown to contain many annotation artifacts, allowing models to shortcut understanding using simple fallible heuristics, and still perform well on the test set.
So it is no surprise that many adversarial (challenge) datasets have been created that cause models trained on standard datasets to fail dramatically.
Although extra training on this data generally improves model performance on just that type of data, transferring that learning to unseen examples is still partial at best.
This work evaluates the failures of state-of-the-art models on existing adversarial datasets that test different linguistic phenomena, and find that even though the models perform similarly on MNLI, they differ greatly in their robustness to these attacks.
In particular, we find syntax-related attacks to be particularly effective across all models, so we provide a fine-grained analysis and comparison of model performance on those examples.
We draw conclusions about the value of model size and multi-task learning (beyond comparing their standard test set performance), and provide suggestions for more effective training data.
\end{abstract}


%
\IEEEpeerreviewmaketitle

\section{Introduction}
In recent years, deep learning models have achieved and continued to improve on state-of-the-art results on many NLP tasks.
However, models that perform extremely well on standard datasets have been shown to be rather brittle and easily tricked.
In particular, the idea of \emph{adversarial} examples or attacks was brought over from computer vision, and various methods of slightly perturbing inputs have been developed that cause models to fail catastrophically \cite{mccoy_right_2019,glockner_breaking_2018,naik_stress_2018}.

Adversarial attacks need to be studied from a security perspective for the deployment of real-world systems, but they are also a powerful lens into \emph{interpretability} of black-box deep learning systems.
By examining the failures of state-of-the-art models, we can learn a lot about what they are really learning, which may give us insights into improving their robustness and general performance.

One philosophical generalization about the cause of failure for all current NLP systems is a lack of deep, `real' understanding of language.
We will focus on the task of natural language inference (NLI), which is a basic natural language understanding task thought to be a key stepping stone to higher-level understanding tasks like question answering and summarization.
The setup of the NLI task is to determine whether a \emph{hypothesis} is true given a \emph{premise}, answering \emph{entailment}, \emph{contradiction}, or \emph{neutral}.

The current top-performing systems for NLI rely on pretraining on generic tasks, followed by fine-tuning on a labeled task-specific dataset.
This is in contrast to older (before late 2018) models, which were primarily task-specific architectures trained on task-specific labeled datasets.
In addition, the Transformer architecture \cite{vaswani_attention_2017} now outperforms the previously dominating recurrent architectures (LSTM and variants).
We want to analyze what kinds of adversarial attacks are still potent on highly-acclaimed recent NLP models like BERT \cite{devlin_bert:_2018} and MT-DNN \cite{liu_multi-task_2019}. 


Our contributions are as follows:
\begin{itemize}
    \item We test models on a variety of existing adversarial datasets, with a high level of granularity to different linguistic phenomena.
    Results indicate that the pre-trained models are remarkably good at lexical meaning, while struggling most with logic and syntactic phenomena.
    \item We focus in on the syntax-focused dataset created by \citeauthor{mccoy_right_2019} \cite{mccoy_right_2019}.
    We look closely at the 30 subcases, and analyze the effects of model size (base vs. large size) and multi-task learning (MT-DNN vs. BERT).
    We also examine what subcases all models fail at.
    \item We experiment with fine-tuning the models with (flattened) dependency parses as input (with no adjustments to architecture or data pre-processing).
    We find that this does improve performance on some, but not all, subcases that rely on the hierarchical structure of sentences.
    \item Lastly, we investigate MNLI's biases by analyzing performance after different amounts of fine-tuning (more and more overfitting) on MNLI.
\end{itemize}


\section{Related Work}
This work joins a growing movement in NLP to go beyond improving test set metrics to more deeply analyze model learning and performance \cite{belinkov_analysis_2019}.
This genre of work believes in the value of interpretability, both to build safer practical systems, and just to find fruitful directions for improving raw model performance.

\citeauthor{liu_inoculation_2019} \cite{liu_inoculation_2019} use a metaphor of inoculation to disentangle the blame for adversarial vulnerability between training data and model architecture.
They expose a small part of the challenge dataset to the model during training, and re-test its evaluation performance on the original test set and the challenge dataset.
\begin{enumerate}
  \item If the model still fails the challenge dataset, the weakness probably lies in its design/architecture or training process.
  \item If the model can now succeed at the challenge dataset (without sacrificing performance on the original dataset), then the original dataset is at fault.
  \item If the model does better on the challenge dataset but worse on the original dataset, the challenge dataset is somehow not representative of the phenomenon it was trying to test, for example having annotation artifacts or being very skewed to a particular label.
\end{enumerate}

Unfortunately, even if adversarial training does improve model performance on that particular dataset, it is fundamentally impossible to devise and train on all possible linguistic phenomena.
The transferability of adversarial robustness to new kinds of examples has been tested by some of the creators of adversarial datasets, by withholding some example generation methods while training on others.
\citeauthor{nie_analyzing_2018} \cite{nie_analyzing_2018} find that knowledge of each of their rule-based templates was almost completely non-transferable to others.
In fact, training on some specific templates caused overfitting and hurt overall robustness.
\citeauthor{mccoy_right_2019} \cite{mccoy_right_2019} find more mixed results, with some cases of successful transfer.

Many standard datasets for different tasks have been shown to have blatant annotation artifacts, allowing models to learn features that are strong in the training (and testing) data, but that have nothing to do with actually performing the task.
\citeauthor{gururangan_annotation_2018} \cite{gururangan_annotation_2018} find many of these artifacts in standard NLI datasets (SNLI and MNLI).
For example, \emph{neutral} hypotheses tend to be longer in length, because an easy way to generate a hypothesis that isn't necessarily entailed by the premise is to add extra details.
Meanwhile, strong negation words like \emph{nobody, no, never} are strong indicators of \emph{contradiction}.
With these artifacts in mind, they split the data into ``hard'' and ``easy'' versions, and model performance decreased by about 15\% on the hard test set.
These findings suggest that it is not the models' faults for failing on adversarial examples, given that there exist easier ways to get high accuracy than truly understanding anything.
But it also means that current evaluation metrics greatly overestimate models' abilities and understanding.



\section{Models}
The two new models that we study gain most of their power from pre-training on a generic language task with a huge unlabeled dataset.
They achieve state-of-the-art performance on a variety of language understanding tasks.
\begin{enumerate}
    \item \textbf{BERT} \cite{devlin_bert:_2018} pre-trains on a bidirectional word-masking language modelling task, in addition to sentence pair prediction, i.e. whether the second sentence is likely to directly follow the first.
    \item\textbf{MT-DNN} \cite{liu_multi-task_2019} builds on BERT by performing multi-task learning on the nine GLUE (General Language Understanding Evaluation) benchmark tasks \cite{wang_glue:_2018}, after BERT's pre-training.
\end{enumerate}
BERT is based on the Transformer architecture \cite{vaswani_attention_2017}, a non-recurrent, purely attention-based architecture.
BERT has a base version (12 Transformer layers), and a large version (24 layers).
We trained base and large versions of both BERT and MT-DNN.
These models are fine-tuned on MNLI starting from publicly available pre-trained checkpoints.

We compare with an older recurrent model, \textbf{ESIM} (Enhanced Sequential Inference  Model) \cite{chen_enhanced_2016}.
It is NLI-task-specific and only trained on MNLI, with no huge pre-training.
It uses a bidrectional LSTM to encode the premise and hypothesis sentences, and uses attention across those representations.
 
 We also considered another model,
    Syntactic TreeLSTM (S-TLSTM), which is identical to ESIM except it uses a TreeLSTM that takes a dependency parse as input \cite{chen_enhanced_2016}.
This model may provide a useful comparison to BERT because its explicit use of the hierarchical structure of language is the exact opposite model design direction from extensive unsupervised pre-training.
However, various studies suggest that the BERT architecture does in fact learn hierarchical structure:
    \citeauthor{goldberg_assessing_2019} \cite{goldberg_assessing_2019} found that BERT performed remarkably well when fine-tuned for external syntactic classification tasks, and \citeauthor{jawahar_what_2019}  \cite{jawahar_what_2019} showed that different layers of BERT learned structural representations of language at different abstraction levels.
\citeauthor{mccoy_right_2019} \cite{mccoy_right_2019} test a different tree-based model (SPINN \cite{bowman_fast_2016}) on their adversarial dataset, and find that it outperforms ESIM, but not BERT.
Considering all this, and the fact that there is currently no tree-based model that comes close to outperforming BERT and variants on standard datasets, we decided not to test S-TLSTM, despite its philosophical appeal.

\section{Overall Results and Analysis}
First, for reference, we provide  the accuracies on the matched MNLI dev set for the models we trained (and tested) in Table \ref{table:overallMNLIResults}. 
$\rm {BERT}_{large}$ results do not quite match published results, but we had limited hardware and did not carefully tune hyperparameters.
The BERT-based models all perform comparably, and even ESIM does respectably.

\begin{table}[!htbp]
\hspace*{1em}
\begin{center}
\begin{tabular}{|l|l|}
\hline
Model & Accuracy (\%)  \\ \hline \hline
ESIM    & 76.80 \\ \hline
BERT base    & 84.17      \\ \hline
BERT large    & \textbf{85.84}      \\ \hline
MT-DNN base    &  84.20     \\ \hline
MT-DNN large    &  \textbf{86.69}     \\ \hline
\end{tabular}
\caption{Overall MNLI Results}
\label{table:overallMNLIResults}
\end{center}
\end{table}

Let us now analyze the performance of the selected models on the adversarial datasets (also called challenge sets, stress tests).
We discuss the first two briefly and then focus on the last one \cite{mccoy_right_2019} because it is the most interesting in terms of actually distinguishing the strengths of the better-performing models. 

\subsection{\textbf{{\citeauthor{glockner_breaking_2018}}} (2018)}
    This dataset is created by modifying SNLI examples with single word replacements of different lexical relations, based on WordNet.
    It tests lexical inferences and relatively simple world knowledge.
    They test a model called KIM (Knowledge-based Inference Model) \cite{chen_enhanced_2016}, which builds on ESIM to explicitly incorporate knowledge from WordNet in a variety of ways, including in architecture additions.
    However, the BERT-based models still significantly outperform KIM.
    This could be due to model architecture, but is most likely a result of their extensive pretraining on a huge diverse corpus.
    There is not a big difference between model sizes, or between MT-DNN and BERT.
    This suggests that lexical semantics is more basic and low-level, so learning it does not need so many layers of abstraction, or multi-task learning (see Table \ref{table:GlocknerAttacks}).

    \begin{table}[!htbp]
      \hspace*{1em}
    \begin{center}
      \begin{tabular}{|l|l|}
        \hline
        Model & Accuracy (\%)  \\ \hline \hline
        ESIM\textsuperscript{*}    & 65.6 \\ \hline
        KIM\textsuperscript{*}    & 83.5 \\ \hline
        BERT base    & 92.2      \\ \hline
        BERT large    & {\bf 94.2}      \\ \hline
        MT-DNN base    &  92.9     \\ \hline
        MT-DNN large    &  {\bf 94.8}     \\ \hline
      \end{tabular}
    \end{center}
      \caption{Single Word Replacement Attacks from \cite{glockner_breaking_2018}. *ESIM and KIM results from original paper.}
      \label{table:GlocknerAttacks}
    \end{table}

\subsection{\textbf{\citeauthor{naik_stress_2018}} (2018)}
    This dataset is composed of a variety of tests motivated by a manual examination and categorization of 100 mistakes made by the best performing model at the time \cite{nie_shortcut-stacked_2017}.
    The categories are antonyms, word overlap (append ``and true is true''), negation words (append ``and false is not true''), length mismatch (append ``and true is true'' 5 times), and spelling errors.
    Antonyms and spelling are ``competence'' tests, while the rest are ``distraction'' tests.
    The examples are generated by modifying examples from MNLI.

    \begin{table}[!htbp]
    \begin{center}
      \begin{tabular}{|l|l|}
      \hline
        Model & Accuracy (\%)  \\ \hline \hline
        ESIM    & 68.39 \\ \hline
        BERT base    & 74.30      \\ \hline
        BERT large    & {\bf 77.21}      \\ \hline
        MT-DNN base    &  73.73     \\ \hline
        MT-DNN large    &  {\bf 77.14}     \\ \hline
      \end{tabular}
      \end{center}
      \caption{Error-analysis motivated attacks from \cite{naik_stress_2018}. Accuracy averaged over all categories of attacks.}
      \label{table:NaikAttacks}
    \end{table}



$\rm {BERT}_{large}$ and $\rm {MT\mbox{-}DNN}_{large}$ do best. 
Overall model performance trends the same as performance on MNLI, but differences are not huge.
Furthermore, when we examined performance on specific categories, all models had about the same pattern of relative performance on different categories of tests, i.e. they have the same relative successes and failures.
This consistency and generally similar performance indicates in this case that the dataset is not well-targeted enough for really interesting insight.
In addition, compared to \citeauthor{mccoy_right_2019} \cite{mccoy_right_2019} (below), the way that examples are generated is more artificial, and maybe less meaningful.
Of course, a robust NLI system still should not be defeated by this kind of attack, i.e. be able to determine irrelevant information, including tautologies, and this test shows that even the best models do not have this capability mastered.


\subsection{\textbf{\citeauthor{mccoy_right_2019}} (2019)}
    They hypothesize that models utilize shallow, fallible syntactic heuristics to achieve accuracy on MNLI, instead of ``real'' understanding.
    The dataset consists of examples generated from manually created templates that break these heuristics.
    They have three categories of heuristics (each is a special case of the one before).
    \begin{enumerate}
        \item Lexical overlap: The model is likely to answer \emph{entailment} if the premise and hypothesis share a lot of words. \\
            It would trick bag-of-words (no word order) models.
        \item Subsequence: The hypothesis is a contiguous string of words from the premise. \\
            \emph{The ball by \underline{the bed rolled}. $\nrightarrow$ The bed rolled.} \\
            It could confuse sequence models too.
        \item Constituent: The hypothesis is a syntactic constituent in the premise. \\
            \emph{If \underline{the boys slept}, they would not eat. $\nrightarrow$ The boys slept.} \\
            It could confuse models that know about syntax.
    \end{enumerate}
    All three heuristics involve the model thinking the answer is \emph{entailment} when it is not, i.e. the \emph{non-entailment} examples are the ones that contradict the heuristic.
    So the extreme imbalance in model performance between entailment and non-entailment examples is strong evidence that the models do indeed rely on the hypothesized heuristics (Table \ref{table:McCoyAttacksEntailment} vs. \ref{table:McCoyAttacksNotEntailment}).


    \begin{table}[!htbp]
    \hspace*{1em}
    \begin{center}
    \begin{tabular}{|l|l|l|l|}
    \hline
    \emph{Entailment}  & word overlap & subseq & constituent \\ \hline
    ESIM        & 96.52           & 98.46       & 94.48       \\ \hline
    $\rm {BERT}_{base}$   & 97.20           & 99.52       & 99.04       \\ \hline
    $\rm {BERT}_{large}$ & {\bf 90.48}           & 99.48       & 96.70       \\ \hline
    $\rm {MT\mbox{-}DNN}_{base}$ & 97.22           & 99.98       & 99.22       \\ \hline
    $\rm {MT\mbox{-}DNN}_{large}$ & 96.06           & 99.54       & 99.14       \\ \hline
    \end{tabular}
    \end{center}
    \caption{Accuracy on examples labeled `entailment'} 
    \label{table:McCoyAttacksEntailment}
    \end{table}

    \begin{table}[!htbp]
    \hspace*{1em}
    \begin{center}
    \begin{tabular}{|l|l|l|l|}
    \hline
    \emph{Non-entailment}  & word overlap & subseq & constituent \\ \hline
    ESIM        & 1.56           & 4.88       & 3.32       \\ \hline
    $\rm {BERT}_{base}$   & 54.68           & 9.46       & 4.88       \\ \hline
    $\rm {BERT}_{large}$  & \textbf{83.44}           & \textbf{31.38}       & \textbf{44.72}       \\ \hline
    $\rm {MT\mbox{-}DNN}_{base}$ & 72.96           & 5.66       & 16.50       \\ \hline
    $\rm {MT\mbox{-}DNN}_{large}$ & \textbf{88.08}           & \textbf{31.24}       & 22.88       \\ \hline
    \end{tabular}
    \end{center}
    \caption{Accuracy on examples labeled `non-entailment'} 
    \label{table:McCoyAttacksNotEntailment}
    \end{table}

All the BERT-based models do significantly better than the LSTM-based ESIM in most categories, as we see in Table \ref{table:McCoyAttacksNotEntailment}. 
But $\rm {BERT}_{large}$ and $\rm {MT\mbox{-}DNN}_{large}$ do vastly better than all others, a difference that was not nearly as apparent in any of the other datasets we tested.
In combination with the granularity in the manually created templates, these huge differences in performance indicate that this dataset more directly probes and reveals the strengths and weaknesses of different models.

The success of $\rm {BERT}_{large}$ and $\rm {MT\mbox{-}DNN}_{large}$ suggests that structural/syntactic information is learned more deeply by a larger model with more layers and parameters to work with (in contrast to lexical semantics (\citeauthor{glockner_breaking_2018}, above)).
$\rm {BERT}_{large}$ also has lower accuracy on the \emph{entailment} examples, also indicating that it is less prone to blindly following the heuristics.

$\rm {MT\mbox{-}DNN}_{base}$ (which is built on $\rm {BERT}_{base}$ and is therefore of comparable size) does significantly better than $\rm {BERT}_{base}$ in some categories, indicating the value of multi-task learning (specifically on language understanding tasks).

\section{Fine-grained Model Comparison}

\begin{table*}[t]
\begin{center}
\begin{tabular}{| p{.6in} | l | p{.5in} | p{.5in} | p{.5in} | p{.5in} | p{.5in} || p{.5in} | p{.5in} | }
\hline
    Heuristic&  Syntactic subcategory&  MT-DNN large & BERT large &  MT-DNN base &   BERT base &   ESIM &  BERT large UP &   MT-DNN base PO  \  \\ \hline \hline
    \multirow{5}{*}{\shortstack{Lexical \\ Overlap}}
    &  subject/object\_swap           & \textbf{0.999} &   \textbf{0.994} &  0.935 &   0.729 &   0     &   0.988          &  0.936 \\ 
    &  preposition                    & 0.934 &  \textbf{0.979} &  0.794 &   0.745 &   0.004 &   0.960          &  0.889 \\ 
    &  relative\_clause               & 0.912 &  \textbf{0.928} &  0.699 &   0.504 &   0.069 &   \textbf{0.930} &  0.837 \\ 
    &  passive                        & \textbf{0.625} &   0.298          &  0.432 &   0.036 &   0     &   0.214          &  0.505 \\ 
    &  conjunction                    & 0.934 &  \textbf{0.973} &  0.788 &   0.720 &   0.005 &   0.943          &  0.711 \\ \hline
    \multirow{5}{*}{Subseq}   
    &  NP/S                           & 0.042 &  0.003          &  0     &   0.016 &   0.058 &   0.004          &  0.003 \\ 
    &  PP\_on\_subject                & 0.668 &  0.673          &  0.168 &   0.293 &   0.001 &   \textbf{0.786} &  0.533 \\ 
    &  relative\_clause\_on\_subject  & 0.749 &  0.698          &  0.082 &   0.133 &   0.087 &   \textbf{0.863} &  0.347 \\ 
    &  past\_participle               & 0.006 &  0.049          &  0.013 &   0.018 &   0.050 &   0.032          &  0.008 \\ 
    &  NP/Z                           & 0.097 &  0.146          &  0.020 &   0.013 &   0.047 &   \textbf{0.217} &  0.172 \\ \hline
    \multirow{5}{*}{Constituent}  
    &  embedded\_under\_if            & 0.703 &  \textbf{0.987} &  0.369 &   0.767 &   0.137 &   0.907          &  0.387 \\ 
    &  after\_if\_clause              & 0.001 &  0              &  0     &   0     &   0     &   0              &  0.010 \\ 
    &  embedded\_under\_verb          & 0.342 &  \textbf{0.903} &  0.252 &   0.299 &   0     &   0.546          &  0.146 \\ 
    &  disjunction                    & 0.005 &  0              &  0.001 &   0.001 &   0.029 &   0.008          &  0.002 \\ 
    &  adverb                         & 0.093 &  \textbf{0.346} &  0.203 &   0.079 &   0     &   0.083          &  0.036 \\ \hline
\end{tabular}
\end{center}
\caption{Results for \emph{non-entailment} subcases. Each row corresponds to a syntactic phenomenon. BERT large UP: trained on unparsed then parsed; MT DNN-base PO: trained on parsed only}
\label{table:FineGrainedAnalysisResults}
\end{table*}

\subsection{Comparison of $\rm {BERT}_{base}$ and $\rm {BERT}_{large}$}
$\rm {BERT}_{large}$ performs better than or equal to $\rm {BERT}_{base}$ (at worst -1\%) on all fifteen \emph{non-entailment} subcases.
Some templates saw particularly large improvement, such as modifying clauses:
    \begin{itemize}
        \item Relative clauses that modify nouns (+42.4\%)  \\
            \emph{The artists that supported the senators shouted. $\nrightarrow$ The senators shouted.}
        \item Prepositional phrase modifiers (+38\%) \\
            \emph{The managers next to the professors performed. $\nrightarrow$ The professors performed.}

    \end{itemize}
    Understanding modifying clauses requires understanding the mechanics of compositional semantics (probably utilizing some kind of hierarchical syntax), which is a basic but crucial step in language understanding.
    So $\rm {BERT}_{large}$'s performance over $\rm {BERT}_{base}$ on these examples is evidence of significantly deeper understanding.

    Another area of improvement is the lexical meanings of special subclasses of verbs and adverbs.
    \begin{itemize}
        \item Non-truth verbs with clause complements (+60.4\%) \\
            \emph{The tourists \underline{said that} the lawyer saw the secretary. $\nrightarrow$ The lawyer saw the secretary.}\\
            This template uses a variety of verbs, all of which suggest but do not entail their complements.
        \item Modal adverbs (+26.7\%) \\
            \emph{\underline{Maybe} the scientist admired the lawyers. $\nrightarrow$ The scientist admired the lawyers.}
    \end{itemize}

    Similarly, passive voice is a special \emph{syntactic} phenomenon that $\rm {BERT}_{large}$ improves on, but still has trouble with.
    \begin{itemize}
        \item Passive voice (3.6\% $\rightarrow$ 29.8\%) \\
            \emph{The managers were advised by the athlete. $\nrightarrow$ The managers advised the athlete.}
    \end{itemize}

    $\rm {BERT}_{base}$ and $\rm {BERT}_{large}$ were trained (pre-training and fine-tuning) on the same data, so the difference in the richness of their learning must reside only in the doubled number of layers in $\rm {BERT}_{large}$.
    These performance improvements are evidence that more layers are necessary for learning all the different special cases of language.

    There are also some partially learned special cases, such as the meaning of ``if'' and related (logical implication).
    \begin{itemize}
        \item 76.6\% $\rightarrow$ 98.7\%: \emph{\underline{Unless} \underline{the professor danced}, the student waited. $\nrightarrow$ The professor danced.}
        \item both 0\%: \emph{\underline{Unless} the bankers called the professor, \underline{the lawyers shouted}. $\nrightarrow$ The lawyers shouted.}
    \end{itemize}

    Meanwhile, all models fail to understand the logical meaning of disjunction (0-2\%).
    \begin{itemize}
        \item \emph{The actor helped the lawyers, or the managers stopped the author. $\nrightarrow$ The actor helped the lawyers.}
    \end{itemize}
    Logic is a very important component of inference as an understanding task, but understandably difficult for statistical models to learn properly, because it is in some sense not probabilistic, in addition to being dependent on exact meanings of single function words.
    Many traditional inference systems relied primarily on formal logic machinery, and finding a way to incorporate that into new models seems like a promising direction.
    Designing and training neural networks that parse and understand formal, symbolic logic is a pretty well-studied problem \cite{evans_can_2018}, and it is certainly known theoretically that general neural networks can represent arbitrary nonlinear logical relations.
    The difficulty is getting natural language models to actually care enough about logic during training to use it correctly for a specific task.
    Many different approaches have been explored recently, including but not limited to modifying the loss function to encourage logical consistency \cite{minervini_adversarially_2018}, rule distillation in a teacher-student network \cite{hu_harnessing_2016}, and indirect supervision using probabilitic logic \cite{wang_deep_2018}.
    To our knowledge, these have not yet been incorporated into state-of-the-art models, but they show promising results on the baseline models tested, especially in lower-resource scenarios.

    All of these special cases are almost certainly encountered in BERT's huge pre-training corpus, but that unsupervised stage does not necessarily teach the model how to use that information towards performing inference.
    This is why larger and larger pre-training may not be the most effective or at least efficient way to achieve language understanding.

    Some of the subsequence templates are still a struggle for all models, including large BERT and $\rm {MT\mbox{-}DNN}$ (\textless 10\%):
    \begin{itemize}
        \item \emph{\underline{The manager knew the athlete} mentioned the actor $\nrightarrow$ The manager knew the athlete.}
        \item \emph{When \underline{the students fought the secretary} ran. $\nrightarrow$ The students fought the secretary.}
    \end{itemize}

    These templates are in the spirit of \emph{garden path sentences}, where local syntactic ambiguity causes a sequential reading of a sentence to lead to an incorrect interpretation.
    This kind of sentence has been studied extensively in cognitive science, specifically language processing, as human readers are first misled and then must backtrack to reanalyze the composition of the sentence to understand it properly \cite{ferreira_recovery_1991,osterhout_brain_1994}.
    \citeauthor{goldberg_assessing_2019} \cite{goldberg_assessing_2019} shows that BERT performs well on complex subject-verb agreement tasks, even without any fine-tuning, indicating that the pre-trained model already has the ability to correctly parse this kind of sentence.
    So the model somehow knows about syntax but does not know how to use it towards the task of inference, a teaching failure that can only be blamed on the inference-task-specific fine-tuning.
    MNLI probably has a low occurrence of complex syntax, but perhaps more importantly, the complete syntactic information is rarely necessary to perform the task.
    Nevertheless, an ability to utilize challenging syntax is an important generalizable skill, because it indicates deep, principled understanding of language.



\subsection{Comparison of $\rm {BERT}$ and $\rm {MT\mbox{-}DNN}$}
Even though ${\rm MT\mbox{-}DNN}_{large}$ performs better on MNLI than $\rm {BERT}_{large}$, $\rm {BERT}$ beats ${\rm MT\mbox{-}DNN}$ on more subcases in this dataset.
In particular, ${\rm MT\mbox{-}DNN}_{large}$ struggles much more with subcases that test special lexical meanings that prevent entailment (number is difference between ${\rm MT\mbox{-}DNN}_{large}$ and $\rm {BERT}_{large}$):
\begin{enumerate}
  \item conditionals: if, unless, whether or not (28.4\%)
  \item `belief' verbs: believed, thought, hoped (56.1\%)
  \item uncertainty adverbs: hopefully, maybe, probably (25.3\%)
\end{enumerate}
The only subcase that ${\rm MT\mbox{-}DNN}_{large}$ is significantly better at is the passive voice (+32.7\%).

$\rm {MT\mbox{-}DNN}$ is trained starting with a pre-trained $\rm {BERT}$ and then fine-tuning on the 9 language understanding tasks in the GLUE benchmark (before fine-tuning again on MNLI).
So if $\rm {MT\mbox{-}DNN}$ performs worse than a $\rm {BERT}$ model of the same size, this fine-tuning caused it to \emph{forget} some knowledge that it had before.
This would happen if the datasets being fine-tuned on do not explicitly test that knowledge, teaching the model to care less about the information from these words.
Considering that most of the GLUE tasks are not straight NLI tasks, it is somewhat unsurprising that the model forgot how these words affect entailment.

\begin{table*}
\begin{center}
\begin{tabular}{| l |   p{3.9in} |   l |}\hline 
Type & Sentence 1 & Sentence 2\\ \hline \hline
NP/S & The manager knew the tourists supported the author.  & The manager knew the tourists.\\ \hline
NP/Z & Since the judge stopped the author contacted the managers.   & The judge stopped the author. \\ \hline
past\_participle & The scientist presented in the school stopped the artists.   &The scientist presented in the school.\\ \hline
after\_if\_clause & Unless the scientists introduced the presidents, the athletes recommended the senator.  & The athletes recommended the senator.  \\ \hline
\end{tabular}
\end{center}
\caption{Non-entailed cases where $\rm {BERT}_{large}$ and $\rm {MT\mbox{-}DNN}_{large}$ do very poorly: Sentence 1 does not entail Sentence 2.}
\label{table:NonEntailedBERT0Cases}
\end{table*}

\section{Parses as Input}
Considering that syntactic phenomena are one of the models' weaknesses, we conduct an experiment of simply passing the flattened binary parses as the input ``sentences''.
We use the automatically generated parses that come with MNLI and the adversarial dataset.
We test on the dataset from \citeauthor{mccoy_right_2019} \cite{mccoy_right_2019}. \\
We try two fine-tuning regimens:
\begin{enumerate}
    \item Fine tune on original (unparsed) MNLI, then fine-tune again on the same data, parsed (labeled UP in Table \ref{table:FineGrainedAnalysisResults}).
    \item Only fine-tune on parsed MNLI (no other inference-specific fine-tuning) (labeled PO in Table \ref{table:FineGrainedAnalysisResults}).
\end{enumerate}
We find that it is rather difficult to get the different models to train well.
Some had loss that never converged, some got near 0\% on all \emph{non-entailment} subcases.
The only reasonable parsed models are $\rm {BERT}_{large}$ under the first regimen (UP), and $\rm {MT\mbox{-}DNN}_{base}$ under the second (PO).
It is likely that these difficulties could be overcome with some systematic hyperparameter tuning, but we see substantial consistency (in model performance on the adversarial dataset) between the two successes, so do not think it would be very insightful to test more.
But the fact that the models responded so differently to fine-tuning suggests that the models have significantly different `knowledge states' in terms of what they learned about how to solve tasks, i.e. they ended up in different local optima after pre-training.
This idea deserves more analysis, because the whole point of huge pre-training is to learn maximally transferable and general representations of language.
Thus, how to guide models towards these ideal local optima (and away from overfitting) is a very important and difficult question.

The fact that any model is able to learn what to do with parses is already  surprising, given that none of their pre-training is parsed.
Evaluating on the parses of MNLI (matched dev set), $\rm {BERT}_{large}$ achieves 82\% accuracy (compare to 86\% unparsed), and $\rm {MT\mbox{-}DNN}_{base}$ gets 84\% (equal to unparsed).

These are the six subcases that saw a 10\% or greater change in accuracy between parsed and unparsed inputs.
Numbers are percent change from unparsed to parsed ($\rm {BERT}_{large}$, $\rm {MT\mbox{-}DNN}_{base}$).\\
Parsing does better on:
\begin{itemize}
    \item Modifiers on subject \\
        \emph{The managers next to \underline{the professors performed}. $\nrightarrow$ The professors performed.} (+11.3\%, +36.5\%) \\
        \emph{The artists that supported \underline{the senators shouted}. $\nrightarrow$ The senators shouted.} (+16.5\%, +26.5\%)
    \item NP/Z (+7.1\%, +15.2\%) \\
        \emph{Since \underline{the athlete hid the secretaries} introduced the president. $\nrightarrow$ The athlete hid the secretaries.} \\
        The parsed models still only achieve 21.7\% and 17.2\% accuracy, but this is still some improvement. 
    \item Conjunction (+22.2\%, +1.8\% (unparsed $\rm {MT\mbox{-}DNN}_{base}$ already gets 90.8\%)) \\
        \emph{The tourists \underline{and} senators admired the athletes $\rightarrow$ The tourists admired the athletes.} \\
        This is an \emph{entailment} template, so $\rm {BERT}_{large}$'s lower accuracy actually indicates less heuristic reliance, and parsed improvement from 64.4\% $\rightarrow$ 86.6\% really indicates better understanding (while $\rm {MT\mbox{-}DNN}_{base}$'s performance could just be using the heuristic).
\end{itemize}
Parsing does worse on:
\begin{itemize}
    \item Embedded clause under non-truth verb (-35.7\%, -10.6\%) \\
        \emph{The lawyers \underline{believed that} the tourists shouted. $\nrightarrow$ The tourists shouted.}
    \item Adverbs indicating uncertainty (-26.3\%, -16.7\%) \\
        \emph{\underline{Hopefully} the presidents introduced the doctors $\nrightarrow$ The presidents introduced the doctors.}
\end{itemize}

Of this small set of significant changes, it can be said that the parsed inputs helped the model with syntactic, hierarchical examples, and hurt it on specific lexical semantics.
This is a surprisingly intuitive result: the model shifted its focus more to syntax!

However, these are the only subcases that changed significantly, out of 30, suggesting either that the parses don't encode that much useful information, or (more likely) that the fine-tuning didn't teach the model how to use the extra information.
For example, maybe $\rm {BERT}_{large}$ (trained on unparsed then the exact same data parsed) just learned to ignore parentheses.

Furthermore, the subcases which had score close to 0 for the unparsed model basically did not see any improvement. These obstinate cases are given in Table \ref{table:NonEntailedBERT0Cases}.
Most of these cases are tests of syntactic phenomena, so parsed data certainly contains useful information, but again, the fine-tuning is somehow not enough to teach the model how to use it.

We do not think that parsing is necessarily a preprocessing step that should be incorporated into future models/systems, because it takes extra computational and annotated data resources. But this experiment does show that without induced biases, BERT's massive, generic pre-training does not capture some basic rule-like principles.

\section{Overfitting to MNLI}
Models learn and use fallible heuristics only because it works on their training datasets; in other words, they \emph{overfit} to their training data, MNLI.
We analyze this process by evaluating the model after different amounts of fine-tuning on MNLI.
We perform this experiment on ${\rm MT\mbox{-}DNN}_{large}$, the best performer on MNLI, and gauge overfitting by evaluating on the adversarial dataset from \citeauthor{mccoy_right_2019} (non-entailment subcases).
\begin{table}[!htbp]
\begin{center}
  \begin{tabular}{| l | l | l | l | l |}
  \hline
    Epoch \# & 1 & 2 & 3 \\ \hline
    MNLI (matched dev set) & 85.66 & \textbf{86.69} & \textbf{86.59} \\ \hline
    \emph{non-entailment} subcases from \cite{mccoy_right_2019} & 44.09 & \textbf{47.40} & 42.49 \\ \hline
  \end{tabular}
  \caption{Accuracy (\%) for ${\rm MT\mbox{-}DNN}_{large}$ fine-tuned on MNLI for varying numbers of epochs, and then evaluated on the dataset from \citeauthor{mccoy_right_2019}. \cite{mccoy_right_2019}}
  \label{table:OverfittingMNLI}
\end{center}
\end{table}

The ${\rm MT\mbox{-}DNN}_{large}$ model trains very quickly, reaching 1\% away from max dev accuracy after only one epoch of fine-tuning, and decreasing slightly on dev accuracy by the third epoch.
This is a claimed benefit of multi-task learning: the model is more flexible to learning different tasks quickly.

From epoch 2 to 3, MNLI dev performance decreases by only 0.1\%, but according to performance on the adversarial dataset, the model is relying significantly more on heuristics, revealing a more overfit state.
Looking at specific subcases, the epoch-3 model differs by more than 10\% in 6 subcases, split very similarly to what happened with parsed inputs:
\begin{itemize}
  \item Improves at lexical semantics: `belief' verbs (believed, thought) (+11.8\%) and uncertainty adverbs (hopefully, maybe) (+24.3\%)
  \item Gets worse at structural/syntactic phenomena: passive voice (-24.4\%), conjunction (-12.4\%), and subject modifiers (PP (-15.6\%), relative clauses (-19.1\%))
\end{itemize}
Interestingly, the subcases that more MNLI fine-tuning helps are exactly the same as the ones that $\rm {BERT}_{large}$ beats ${\rm MT\mbox{-}DNN}_{large}$ on.
This strongly suggests that the purpose of these words is emphasized in MNLI; ${\rm MT\mbox{-}DNN}$ forgets about it while fine-tuning on other GLUE tasks, and more fine-tuning on MNLI makes it re-learn it.

On the other hand, the subcases that more fine-tuning hurts are all structural/syntax-focused, indicating that MNLI is biased against actually utilizing complex syntactic phenomena in a way that affects entailment (supporting the \emph{syntactic} heuristic hypothesis of \citeauthor{mccoy_right_2019}).

Creating feasibly-sized training datasets with ``no biases'' is impossible.
Here we find some subtle examples in MNLI, emphasizing the sensitivity of these models to pick up on any useful signal.
NLI is a very broad task, making it hard to define what a natural or representative input distribution would be, so ultimately dataset design should depend on desired abilities and applications.

\section{Conclusion}
In this work, we use adversarial and challenge datasets to probe and analyze the failures of current state-of-the-art natural language inference models, comparing BERT and MT-DNN models of different sizes.
Evaluating on these datasets distinguishes the actual understanding capabilities of the different models better than simply looking at their performance on MNLI (the large dataset they were trained on).
Our analysis is very fine-grained, targeting many specific linguistic phenomena.
We find various improvements from larger model size and multi-task learning.
We find that the most difficult examples for the best models are logic or syntax-based, including propositional logic and garden-path sentences.
We experiment with passing parses as input to the out-of-the-box pre-trained models, and find that it does provide some improvement in examples that require understanding syntax, demonstrating the value of syntactic induced biases.
We analyze what overfitting to MNLI looks like, and reveal some biases/artifacts in the dataset.

Some may argue that testing NLI systems on artificially challenging datasets is unfair and not useful, because it is not representative of their performance on naturalistic, real-world data.
But even if the data humans naturally produce is not so difficult (because humans also are lazy and use heuristics), the difference is that we always \emph{can} parse sentences correctly, utilizing rules and principles.
And we intuitively know that ability is crucial to robust, trustworthy, and \emph{real} language understanding.


\section*{Acknowledgment}
The work reported in this paper is supported by the National Science Foundation under Grant No. 1659788. Any opinions, findings and conclusions or recommendations expressed in this work are those of the author(s) and do not necessarily reflect the views of the National Science Foundation.

\balance



\bibliographystyle{IEEEtranN}
\bibliography{MyLibrary}
%



\end{document}